# A Novel Differentiable Loss Function for Unsupervised Graph Neural Networks in Graph Partitioning

Vivek Chaudhary [0000-0002-5517-3190]

`vivekch2018@gmail.com`

**Abstract.** In this paper, we explore the graph partitioning problem, a pivotal combinatorial optimization challenge with extensive applications in various fields such as science, technology, and business. Recognized as an NP-hard problem, graph partitioning lacks polynomial-time algorithms for its resolution. Recently, there has been a burgeoning interest in leveraging machine learning, particularly approaches like supervised, unsupervised, and reinforcement learning, to tackle such NP-hard problems. However, these methods face significant hurdles: supervised learning is constrained by the necessity of labeled solution instances, which are often computationally impractical to obtain; reinforcement learning grapples with instability in the learning process; and unsupervised learning contends with the absence of a differentiable loss function, a consequence of the discrete nature of most combinatorial optimization problems. Addressing these challenges, our research introduces a novel pipeline employing an unsupervised graph neural network to solve the graph partitioning problem. The core innovation of this study is the formulation of a differentiable loss function tailored for this purpose. We rigorously evaluate our methodology against contemporary state-of-the-art techniques, focusing on metrics: cuts and balance, and our findings reveal that our is competitive with these leading methods.

## 1  Introduction

In recent years, there has been increasing interest in the application of machine learning for solving combinatorial optimization problems. These problems, inherently complex and computationally challenging, span a wide range of applications, from the well-known Travelling Salesman Problem [1] to the Knapsack Problem [2]. Among these, notable examples include the Maximum Clique Problem [3], Maximum/Minimum Cut Problem [4], Boolean Satisfiability Problem [5], Hamiltonian Path Problem, Facility Location Problem [6], and the Quadratic Assignment Problem [7]. The diversity and complexity of these problems pose unique challenges and opportunities for machine learning approaches.

Machine learning methodologies employed to tackle these problems can be broadly categorized into supervised learning, unsupervised learning, and reinforcement learning. Each approach offers distinct perspectives and tools for solving these computationally intensive problems. Supervised learning, despite its widespread use [12] [13] [14], encounters a fundamental challenge in this domain: it requires large datasets of pre-solved instances, leading to a 'chicken and egg' situation [8]. Another problem that complicates supervised learning is the difficulty in sampling unbiased labeled instances of NP-hard problems [3]   .

Reinforcement learning, while showing promise in discrete action spaces [9] [11], faces hurdles due to its lack of full differentiability, making training procedures complex and resource intensive [10]. This limitation is especially pronounced when dealing with the vast and intricate search spaces of NP-hard problems.

Unsupervised learning, on the other hand, has recently become a focal point in the machine learning community [15] [16], especially given its potential to circumvent the need for labeled data. The primary challenge here lies in developing an effective loss function that can guide the learning process in the absence of labeled examples.

Graph Neural Networks (GNNs) have been a significant innovation within the deep learning community, particularly for data with graph structures. GNNs excel at learning feature representations for nodes, edges, and entire graphs, making them well-suited for a variety of applications [17]. These applications range from user classification in social networks [19] to predicting interactions in recommender systems [18] and properties of molecular graphs [20]. The versatility and effectiveness of GNNs in modeling complex structural data have led to their successful application across a broad spectrum of real-world problems.

Our research aligns with the cutting-edge efforts in the deep learning community, especially those focusing on training unsupervised GNNs end to end. We leverage the potential of GNNs to tackle the graph partitioning problem, a classic example of an NP-hard problem, without the reliance on labeled training sets. This approach not only aligns with the recent trends in machine learning but also addresses the specific challenges posed by NP-hard combinatorial optimization problems. By doing so, our work seeks to contribute to the growing body of knowledge in applying unsupervised learning techniques, particularly GNNs, to complex, real-world optimization challenges. Section 2 outlines theoretical preliminaries for this paper. Section 3 contains details of the probabilistic graph neural network for graph partitioning. Section 4 contains the details and results of our experiments. The work is concluded in section 5.

## 2 Preliminaries

### 2.1 Graph partitioning problem

A graph $G=(V,E)$ consists of a set of vertices $V$ and a set of edges $E$. The objective in graph partitioning is to divide the vertex set $V$ into disjoint subsets $V_1$, $V_2$, such that certain conditions or objectives are satisfied. The objective is to minimize the number of edges between different subsets while maintaining a balance in the size of these subsets [32].

Proving the NP-hardness of graph partitioning involves demonstrating that no polynomial-time algorithm can solve it for all cases unless $P=NP$. This complexity leads to a focus on approximation algorithms, which seek near-optimal solutions within a reasonable time frame. Performance of these algorithms is often evaluated based on how close they get to the optimal solution. State of the Art methods for solving graph partitioning include Kernighan-Lin method [29], Spectral Partitioning [30], and Multi-Level Partitioning [31].

Besides minimizing the edge cut and maintaining balance, other metrics like modularity (for community detection) [33], communication volume [34] (in parallel computing), and expansion or conductance (measuring how well connected the subgraphs are) can be crucial, depending on the application. In this paper, we focus on percent of cuts and balance.
.

### 2.2 Graph Neural Networks

Graph Neural Networks (GNNs) represent a significant advancement in the field of deep learning, particularly for processing data that is naturally structured as graphs. Unlike standard neural networks that assume a sequential or grid-like structure in the input data, GNNs are designed to work directly with graph structures, consisting of nodes (vertices) and edges. This section provides a detailed technical and mathematical exposition of GNNs, elucidating their fundamental concepts, and architecture.

**Fundamental Concepts**

*Graph Structure.* A graph $G=(V,E)$ consists of a set $V$ containing vertices and a set $E$ containing edges. Each vertex $v \in V$ can have associated features $x_v$, and similarly, edges can have features $x_e$.

*Node representation.* The core objective of a GNN is to learn a representation (embedding) $h_v$ for each node $V$, which captures both its features and its structural role within the graph.

**Mathematical Framework.** The core operation of a GNN is the aggregation of information from a node's neighborhood. This can be formalized as follows:

*Message Passing:* Each node $v$ aggregates messages from adjacent nodes $N(v)$ and possibly its own features. This is typically a two-step process involving message calculation and aggregation [28]:

1. Message Calculation: For each edge $(u, v) \in E$, a message function M computes a message $m_{uv} = M(h_u, h_v, x_e)$, where $h_u$ and $h_v$ are the feature vectors of the nodes and $x_e$ is the edge feature.
2. Aggregation: A node v aggregates messages from its neighborhood using an aggregation function $A$, $a_v = A(\{m_{uv} : u \in N(v)\})$. Common aggregation functions include sum, mean, and max.

*Node Update.* The aggregated message $a_v$ is then combined with the node's current state to update its representation. This update is typically done using a neural network $U: h'_v = U(h_v, a_v)$. After this step, $h'_v$ becomes the new feature representation of node $v$.

*Learning Objective.* The learning process in GNNs involves adjusting the parameters of the message, aggregation, and update functions to optimize a task-specific objective, such as node classification [29], graph classification [30], or link prediction [31].

**Architecture Variants**

*Graph Convolutional Networks (GCNs):* These generalize convolutional neural networks to graphs by defining convolution operations on the graph structure [25].

*Graph Attention Networks (GATs):* GATs apply attention mechanisms in the aggregation step, allowing the model to learn the importance of each neighbor's message [26].

*GraphSAGE:.* This variant samples a fixed-size neighborhood and uses different aggregation functions, enabling scalability to large graphs [27].

## 3   Probabilistic Graph Neural Network for Graph Partitioning

We assume a weighted graph $G = (V,E,w)$ where $V$ is the set of edges, $E$ is the set of edges and w is the set of weights of the nodes. The graph partitioning can be modelled as

$$\min_{S \subseteq v} k(S; G) \ such\ that\ S \in \Upsilon \tag{1}$$

Where k is a cost function, Y is the family of sets which conform to a partitioned graph.

### 3.1 Solution pipeline

Our solution is inspired from [3] and comprises of the following steps. A GNN $n_\theta$ is constructed that outputs a distribution $D = n_\theta(G)$ over sets. GNN $n\theta$ is trained to optimize the probability that there is a valid $S^* \sim D$ with a small cost $k(S*; G)$. $S^*$ is recovered from $D$ deterministically.

### 3.2 Probabilistic loss function for Graph Partitioning Problem

We train the model to obtain a distribution with a low cost We define a probabilistic loss function $l(D;G)$ that adheres to the following.

$$P(k(S;G) < l(D;G) > z \quad with\ D = n_\theta(G) \tag{2}$$

By using Markov's inequality, we get:

$$l(D;G) \triangleq \frac{E[k(S;G)]}{1-z} \quad for\ any\ z \in [0,1) \tag{3}$$

Now if we train the GNN to a sufficiently small loss $l(D;G) = \varepsilon$, then there exists a positive probability that a set $S^*$ exists for which the loss is at most $\varepsilon$. For the graph partitioning problem, the loss function can be described as

$$l(D;G) \triangleq l_{cuts} + l_{balance} + l_{centrality} \tag{4}$$

Where $l_{cuts}$ is the loss for a cut, or an edge between two different partitions. This term is given by:

$$l_{cuts} = \Sigma((\tan(\alpha * m_i) - 0.5) * (\tan(\alpha * m_j) - 0.5)) \tag{5}$$

Where $m_i$ and $m_j$ are estimations of index of maximum arguments in SoftMax outputs $p_i$ and $p_j$ for adjacent nodes. α is a parameter that can be tuned while training. $l_{balance}$ is the loss for imbalance, or size difference in the two partitions. This term is given by:

$$l_{bala} = \Sigma\tan(\alpha * (f_i - 0.5))^2 + \Sigma\tan(\alpha * (f_j - 0.5))^2 \tag{6}$$

Where $f_i$ and $f_j$ are the SoftMax outputs for the first partition in SoftMax outputs $p_i$ and $p_j$ for adjacent nodes.

$l_{centrality}$ is a loss for ensuring that the SoftMax outputs $p_i$ and $p_j$ are not centered around 0.5. This term is given by:

$$l_{centrality} = e^{-\frac{(f_i + f_j)^2}{2*\xi^2}} \quad (7)$$

Where ξ is a parameter that can be tuned while training.

## 4 Experiment results

We conduct experiments on sets of graphs with different number of nodes. The performance of the models is measured in terms of cut percentage and imbalance percentage. Cut percentage is the percentage of number of edges that violate the constraints to the total number of edges. Imbalance percentage is the percentage of difference in size of partitions to the total number of nodes. Our Probabilistic GNN is compared with Kernighan-Lin method and Spectral Partitioning.

**Table 1.** Cut percentage for different methods for different graph sizes

| Nodes | GNN | Kernighan-Lin | Spectral |
| --- | --- | --- | --- |
| 50 | 33.50 | 31.69 | 26.82 |
| 100 | 38.42 | 36.53 | 32.33 |
| 150 | 40.50 | 38.84 | 34.96 |
| 200 | 41.62 | 40.19 | 37.06 |
| 250 | 42.80 | 41.24 | 35.44 |
| 300 | 43.15 | 41.95 | 36.46 |
| 350 | 45.57 | 42.52 | 37.04 |
| 400 | 44.29 | 42.99 | 37.10 |
| 450 | 44.47 | 43.39 | 37.17 |
| 500 | 44.50 | 43.69 | 38.93 |

**Table 2.** Imbalance percent for different methods for different graph sizes

| Nodes | GNN | Kernighan-Lin | Spectral |
|---|---|---|---|
| 50 | 6.80 | 0.00 | 42.56 |
| 100 | 5.48 | 0.00 | 40.96 |
| 150 | 5.28 | 0.00 | 38.96 |
| 200 | 5.30 | 0.00 | 36.70 |
| 250 | 3.36 | 0.00 | 44.27 |
| 300 | 3.32 | 0.00 | 42.59 |
| 350 | 7.63 | 0.00 | 41.73 |
| 400 | 2.72 | 0.00 | 42.62 |
| 450 | 4.63 | 0.00 | 42.28 |
| 500 | 6.73 | 0.00 | 38.54 |

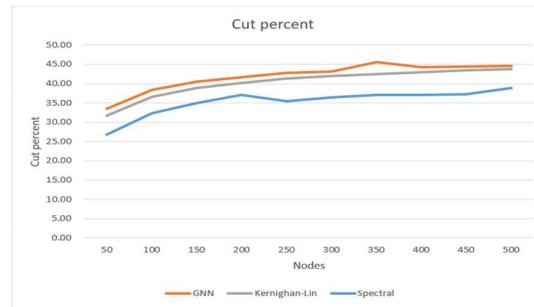

**Fig. 1.** Cut percent for different methods

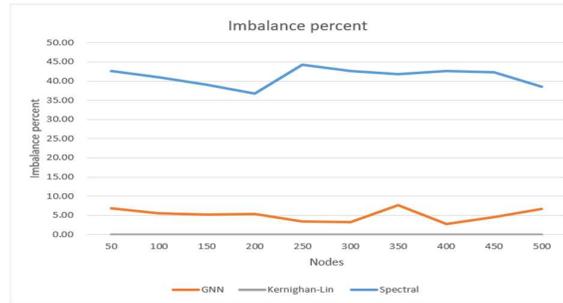

**Fig. 2.** Imbalance percent for different methods

From the above observations, it is evident that our method is competitive with Kernighan-Lin method in terms of cut percentage, lying within 10% of its value. Our method performs better than Spectral method when it comes to imbalance percentage. Our method produces more balanced graphs than Spectral method, though with more cuts.

## 5      Conclusion

In this paper we proposed a novel method for solving Graph Partitioning Problem with Probabilistic Graph Neural Network. With our experimentation results we demonstrated that our novel method is competitive with state-of-the-art methods in certain instances.

## References

1. Yong Shi, Yuanying Zhang, The neural network methods for solving Traveling Salesman Problem, Procedia Computer Science, Volume 199, 2022, Pages 681-686, ISSN 1877-0509, https://doi.org/10.1016/j.procs.2022.01.084.
2. Fehmi Burcin Ozsoydan, İlker Gölcük, A reinforcement learning based computational intelligence approach for binary optimization problems: The case of the set-union knapsack problem, Engineering Applications of Artificial Intelligence, Volume 118, 2023, 105688, ISSN 0952-1976, https://doi.org/10.1016/j.engappai.2022.105688.
3. Karalias, Nikolaos, and Andreas Loukas. "Erdos goes neural: an unsupervised learning framework for combinatorial optimization on graphs." Advances in Neural Information Processing Systems 33 (2020): 6659-6672.
4. Melnikov, Alexey, et al. "Quantum machine learning: From physics to software engineering." Advances in Physics: X 8.1 (2023): 2165452.
5. Guo, Wenxuan, et al. "Machine learning methods in solving the boolean satisfiability problem." Machine Intelligence Research (2023): 1-16.
6. Agrawal, Priyank, et al. "Learning-augmented mechanism design: Leveraging predictions for facility location." Proceedings of the 23rd ACM Conference on Economics and Computation. 2022.
7. Wang, Runzhong, Junchi Yan, and Xiaokang Yang. "Neural graph matching network: Learning lawler's quadratic assignment problem with extension to hypergraph and multiple-graph matching." IEEE Transactions on Pattern Analysis and Machine Intelligence 44.9 (2021): 5261-5279.
8. Yehuda, Gal, Moshe Gabel, and Assaf Schuster. "It's not what machines can learn, it's what we cannot teach." International conference on machine learning. PMLR, 2020.
9. Vinyals, Oriol, et al. "Grandmaster level in StarCraft II using multi-agent reinforcement learning." Nature 575.7782 (2019): 350-354.
10. Thrun, Sebastian, and Anton Schwartz. "Issues in using function approximation for reinforcement learning." Proceedings of the 1993 connectionist models summer school. Psychology Press, 2014.
11. Silver, David, et al. "A general reinforcement learning algorithm that masters chess, shogi, and Go through self-play." Science 362.6419 (2018): 1140-1144.
12. Joshi, Chaitanya K., Thomas Laurent, and Xavier Bresson. "An efficient graph convolutional network technique for the travelling salesman problem." arXiv preprint arXiv:1906.01227 (2019).
13. Li, Zhuwen, Qifeng Chen, and Vladlen Koltun. "Combinatorial optimization with graph convolutional networks and guided tree search." Advances in neural information processing systems 31 (2018).

14. Lemos, Henrique, et al. "Graph colouring meets deep learning: Effective graph neural network models for combinatorial problems." 2019 IEEE 31st International Conference on Tools with Artificial Intelligence (ICTAI). IEEE, 2019.
15. Li, Wei, et al. "Rethinking Graph Neural Networks for Graph Coloring." (2020).
16. Schuetz, Martin JA, J. Kyle Brubaker, and Helmut G. Katzgraber. "Combinatorial optimization with physics-inspired graph neural networks." Nature Machine Intelligence 4.4 (2022): 367-377.
17. Hamilton, William L. Graph representation learning. Morgan & Claypool Publishers, 2020.
18. Fan, Wenqi, et al. "Graph neural networks for social recommendation." The world wide web conference. 2019.
19. Song, Chenguang, Kai Shu, and Bin Wu. "Temporally evolving graph neural network for fake news detection." Information Processing & Management 58.6 (2021): 102712.
20. Wieder, Oliver, et al. "A compact review of molecular property prediction with graph neural networks." Drug Discovery Today: Technologies 37 (2020): 1-12.
21. Wu, Felix, et al. "Simplifying graph convolutional networks." International conference on machine learning. PMLR, 2019.
22. Veličković, Petar, et al. "Graph attention networks." arXiv preprint arXiv:1710.10903 (2017).
23. Hamilton, Will, Zhitao Ying, and Jure Leskovec. "Inductive representation learning on large graphs." Advances in neural information processing systems 30 (2017).
24. Scarselli, Franco, et al. "The graph neural network model." IEEE transactions on neural networks 20.1 (2008): 61-80.
25. Xiao, Shunxin, et al. "Graph neural networks in node classification: survey and evaluation." Machine Vision and Applications 33 (2022): 1-19.
26. Errica, Federico, et al. "A fair comparison of graph neural networks for graph classification." arXiv preprint arXiv:1912.09893 (2019).
27. Zhang, Muhan, and Yixin Chen. "Link prediction based on graph neural networks." Advances in neural information processing systems 31 (2018).
28. Brélaz, Daniel. "New methods to color the vertices of a graph." Communications of the ACM 22.4 (1979): 251-256.
29. Kernighan, Brian W., and Shen Lin. "An efficient heuristic procedure for partitioning graphs." The Bell system technical journal 49.2 (1970): 291-307.
30. McSherry, Frank. "Spectral partitioning of random graphs." Proceedings 42nd IEEE Symposium on Foundations of Computer Science. IEEE, 2001.
31. Hendrickson, Bruce, and Robert W. Leland. "A Multi-Level Algorithm For Partitioning Graphs." SC 95.28 (1995): 1-14.
32. Andreev, Konstantin, and Harald Räcke. "Balanced graph partitioning." Proceedings of the sixteenth annual ACM symposium on Parallelism in algorithms and architectures. 2004.
33. Newman, Mark EJ. "Modularity and community structure in networks." Proceedings of the national academy of sciences 103.23 (2006): 8577-8582.
34. Hendrickson, Bruce, and Tamara G. Kolda. "Graph partitioning models for parallel computing." Parallel computing 26.12 (2000): 1519-1534.